\documentclass[runningheads]{llncs}
\usepackage{graphicx}
\usepackage{comment}
\usepackage{amsmath,amssymb} 
\usepackage{color}
\usepackage{xcolor}
\usepackage{algorithm}
\usepackage{algorithmic}
\usepackage{multirow}
\usepackage{array}
\usepackage{hyperref}

\begin{document}
\pagestyle{headings}
\mainmatter
\def\ECCVSubNumber{700}  
\def\etal{\textit{et al.}}

\title{Prototype Mixture Models \\for Few-shot Semantic Segmentation} 

\titlerunning{Prototype Mixture Models}
%
\author{Boyu Yang \and
Chang Liu \and
Bohao Li \and
Jianbin Jiao \and
Qixiang Ye*}
\authorrunning{Boyu Yang et al.}
%
\institute{University of Chinese Academy of Sciences, Beijing, China\\
\email{\{yangboyu18, liuchang615\}@mails.ucas.ac.cn, libohao1998@gmail.com}\\
\email{\{jiaojb, qxye\}@ucas.ac.cn}}
\maketitle

\begin{abstract}
Few-shot segmentation is challenging because objects within the support and query images could significantly differ in appearance and pose. Using a single prototype acquired directly from the support image to segment the query image causes semantic ambiguity. In this paper, we propose prototype mixture models (PMMs), which correlate diverse image regions with multiple prototypes to enforce the prototype-based semantic representation. Estimated by an Expectation-Maximization algorithm, PMMs incorporate rich channel-wised and spatial semantics from limited support images. Utilized as representations as well as classifiers, PMMs fully leverage the semantics to activate objects in the query image while depressing background regions in a duplex manner. Extensive experiments on Pascal VOC and MS-COCO datasets show that PMMs significantly improve upon state-of-the-arts. Particularly, PMMs improve 5-shot segmentation performance on MS-COCO by up to 5.82\% with only a moderate cost for model size and inference speed. 
\footnote{Code is available at \href{https://github.com/Yang-Bob/PMMs}{\color{magenta}github.com/Yang-Bob/PMMs}. \\ *Qixiang Ye is the corresponding author.}

\end{abstract}


\begin{figure}[t]
\centering
\includegraphics[width=1\columnwidth]{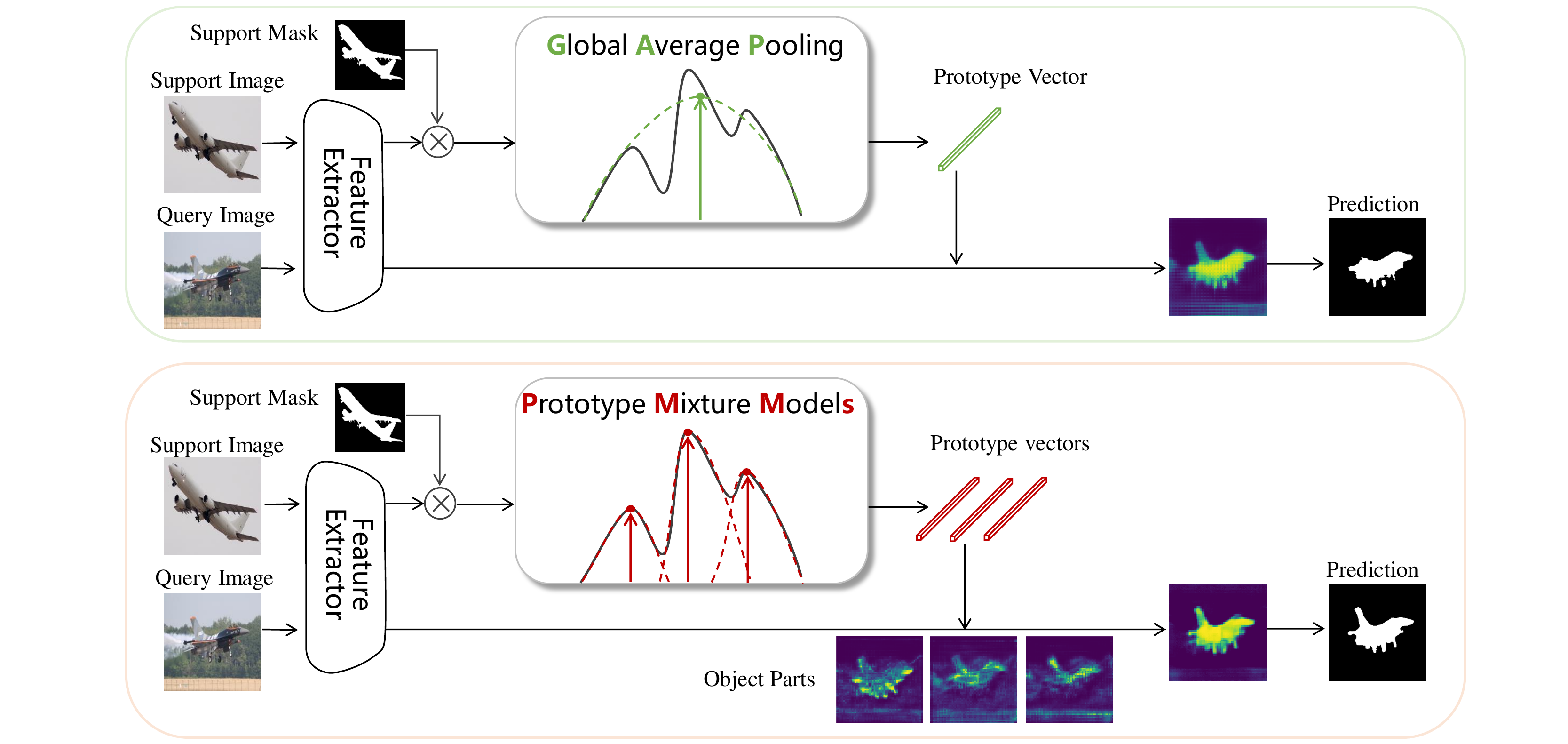}
\caption{The single prototype model (upper) based on global average pooling causes semantic ambiguity about object parts. In contrast, prototype mixture models (lower) correlate diverse image regions, $e.g.$, object parts, with multiple prototypes to enhance few-shot segmentation model.  (Best viewed in color)}
\label{fig:motivation}
\vspace{-0.2cm}
\end{figure}

\section{Introduction}

Substantial progress has been made in semantic segmentation~\cite{PSPNet,UNet,DeepLabV1,DeepLabV2,DeepLabV3,DeepLabV3+,SegSort,MaskRCNN-PAMI2020,EMA}. This has been broadly attributed to the availability of large datasets with mask annotations and convolutional neural networks (CNNs) capable of absorbing the annotation information. However, annotating object masks for large-scale datasets is laborious, expensive, and can be impractical~\cite{CMIL2019,MinEntropy2019,FreeAnchor2019}. It is also not consistent with cognitive learning, which can build a model upon few-shot supervision~\cite{CompositionalRepre-ICCV2019}.

Given a few examples, termed \textit{support} images, and the related segmentation masks~\cite{FWB-ICCV2019}, few-shot segmentation aims to segment the \textit{query} images based on a feature representation learned on training images. It remains a challenging problem when we consider that target category is not included in the training data while objects within the support and query image significantly differ in appearance and pose.

By introducing the metric learning framework, Shaban~\etal~\cite{OSLSM}, Zhang \etal~\cite{SG-One}, and Dong \etal~\cite{BMVC18Prototype} contributed early few-shot semantic segmentation methods. They also introduced the concept of ``prototype" which refers to a weight vector calculated with global average pooling guided by ground-truth masks embedded in feature maps. Such a vector squeezing discriminative information across feature channels is used to guide the feature comparison between support image(s) and query images for semantic segmentation. 

Despite clear progress, we argue that the commonly used prototype model is problematic when the spatial layout of objects is completely dropped by global average pooling, Fig.\ \ref{fig:motivation}(upper). A single prototype causes semantic ambiguity around various object parts and deteriorates the distribution of features~\cite{CollectSelect19}. Recent approaches have alleviated this issue by prototype alignment~\cite{PANet}, feature boosting~\cite{FWB-ICCV2019}, and iterative mask refinement~\cite{CaNet}. However, the semantic ambiguity problem caused by global average pooling remains unsolved.

In this paper, we propose prototype mixture models (PMMs) and focus on solving the semantic ambiguity problem in a systematic manner.
During the training procedure, the prototypes are estimated using an Expectation-Maximization (EM) algorithm, which treats each deep pixel (a feature vector) within the mask region as a positive sample.  
PMMs are primarily concerned with representing the diverse foreground regions by estimating mixed prototypes for various object parts, Fig.\ \ref{fig:motivation}(lower). They also enhance the discriminative capacity of features by modeling background regions.

\begin{figure}[t]
\centering
\includegraphics[width=1.0\columnwidth]{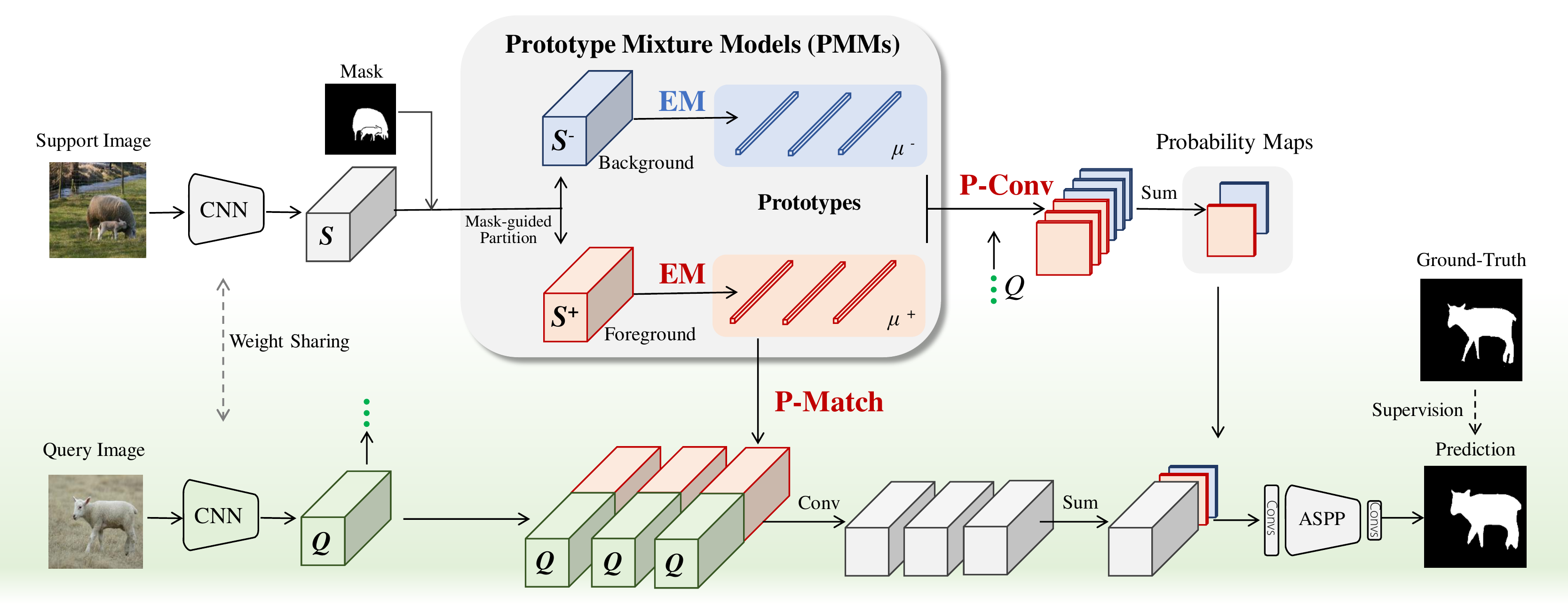}
\caption{The proposed approach consists of two branches $i.e.,$ the support branch and the query branch. During training, the feature set $S$ of a support image is partitioned into a positive sample set $S^+$ and a negative sample set $S^-$ guided by the ground-truth mask. $S^+$ and $S^-$ are respectively used to train $\mu^+$ and $\mu^-$, which are used to activate query features in a duplex way (P-Conv and P-Match) for semantic segmentation. 
``ASPP" refers to the Atrous Spatial Pyramid Pooling (ASPP)~\cite{DeepLabV2}.}
\label{fig:flowchart}
\vspace{-0.2cm}
\end{figure}

The few-shot segmentation procedure is implemented in a metric learning framework with two network branches (a support branch and a query branch), Fig.\ \ref{fig:flowchart}. In the framework, PMMs are utilized in a duplex manner to segment a query image. 
On the one hand, they are regarded as spatially squeezed representation, which match (P-Match) with query features to activate feature channels related to the object class. On the other hand, each vector is regarded as a $C$-dimensional linear classifier, which multiplies (P-Conv) with the query features in an element-wised manner to produce a probability map. In this way, the channel-wised and spatial semantic information of PMMs is fully explored to segment the query image.

The contributions of our work are summarized as follows:
\begin{itemize}
    \item We propose prototype mixture models (PMMs), with the target to enhance few-shot semantic segmentation by fully leveraging semantics of limited support image(s). PMMs are estimated using an Expectation-Maximization (EM) algorithm, which is integrated with feature learning by a plug-and-play manner. 
    
    \item We propose a duplex strategy, which treats PMMs as both representations and classifiers, to activate spatial and channel-wised semantics for segmentation. 
    
    \item We assemble PMMs to RPMMs using a residual structure and significantly improve upon the state-of-the-arts.
\end{itemize}

\section{Related work}

\textbf{Semantic Segmentation.}
Semantic segmentation, which performs per-pixel classification of a class of objects, has been extensively investigated. State-of-the-art methods, such as UNet~\cite{UNet}, PSPNet~\cite{PSPNet}, DeepLab~\cite{DeepLabV1,DeepLabV2,DeepLabV3}, are based on fully convolutional networks (FCNs)~\cite{FCN}. Semantic segmentation has been updated to instance segmentation~\cite{MaskRCNN-PAMI2020} and panoptic segmentation ~\cite{PanopticFPN-CVPR2019}, which shared useful modules, $e.g.$, Atrous Spatial Pyramid Pooling (ASPP)~\cite{DeepLabV2} and multi-scale feature aggregation~\cite{PSPNet}, with few-shot segmentation. The clustering method used in SegSort~\cite{SegSort}, which partitioned objects into parts using a divide-and-conquer strategy, provides an insight for this study.

\textbf{Few-shot Learning.}
Existing methods can be broadly categorized as either: metric learning~\cite{MatchNetwork16,Compare2018,CollectSelect19}, meta-learning ~\cite{LearningToLearn16,Optimization17,MAML17,TaskAgnosticMeta19}, or data argumentation.
Metric learning based methods train networks to predict whether two images/regions belong to the same category. Meta-learning based approaches specify optimization or loss functions which force faster adaptation of the parameters to new categories with few examples. The data argumentation methods learn to generate additional examples for unseen categories~\cite{Hallucinating17,Imaginary18}. 

In the metric learning framework, the effect of prototypes for few-shot learning has been demonstrated. With a simple prototype, $e.g.$, a linear layer learned on top of a frozen CNN~\cite{CloserLook19}, state-of-the-art results can be achieved based on a simple baseline. This provides reason for applying prototypes to capture representative and discriminative features.

\textbf{Few-shot Segmentation. }
Existing few-shot segmentation approaches largely followed the metric learning framework, $e.g.$, learning knowledge using a prototype vector, from a set of support images, and then feed learned knowledge to a metric module to segment query images~\cite{PANet}.

In OSLSM~\cite{OSLSM}, a two-branch network consisting of a support branch and a query branch was proposed for few-shot segmentation. The support branch is devoted to generating a model from the support set, which is then used to tune the segmentation process of an image in the query branch.
In PL~\cite{BMVC18Prototype}, the idea of prototypical networks was employed to tackle few-shot segmentation using metric learning.
SG-One~\cite{SG-One} also used a prototype vector to guide semantic segmentation procedure. To obtain the squeezed representation of the support image, a masked average pooling strategy is designed to produce the prototype vector. A cosine similarity metric is then applied to build the relationship between the guidance features and features of pixels from the query image. 
PANet~\cite{PANet} further introduced a prototype alignment regularization between support and query branches to fully exploit knowledge from support images for better generalization.
CANet ~\cite{CaNet} introduced a dense comparison module, which effectively exploits multiple levels of feature discriminativeness from CNNs to make dense feature comparison. With this approach comes an iterative optimization module which reﬁnes segmentation masks.
%
The FWB approach~\cite{FWB-ICCV2019} focused on discriminativeness of prototype vectors (support features) by leveraging foreground-background feature differences of support images. It also used an ensemble of prototypes and similarity maps to handle the diversity of object appearances.

As a core of metric learning in few-shot segmentation, the prototype vector was commonly calculated by global average pooling. 
However, such a strategy typically disregards the spatial extent of objects, which tends to mix semantics from various parts. This unintended mixing seriously deteriorates the diversity of prototype vectors and feature representation capacity. 
Recent approaches alleviated this problem using iterative mask refinement~\cite{CaNet} or model ensemble~\cite{FWB-ICCV2019}. 
However, issues remain when using single prototypes to represent object regions and the semantic ambiguity problem remains unsolved. 

Our research is inspired by the prototypical network~\cite{PrototypicalNet17}, which learns a metric space where classification is performed using distances to the prototype of each class. The essential differences are twofold: (1) A prototype in prototypical network~\cite{PrototypicalNet17} represents a class of samples while a prototype in our approach represents an object part; (2) The prototypical network does not involve mixing prototypes for a single sample or a class of samples.

\section{The Proposed Approach}

\subsection{Overview}
The few-shot segmentation task is to classify pixels in query images into foreground objects or backgrounds by solely referring to few labeled support images containing objects of the same categories. 
The goal of the training procedure is to learn a segmentation model that is trained by numbers of images different from the task query image categories.
The training image set is split into many small subsets and within every subset one image serves as the query and the other(s) as the support image(s) with known ground-truth(s).
Once the model is trained, the segmentation model is fixed and requires no optimization when tested on a new dataset~\cite{CaNet}.
The proposed few-shot segmentation model follows a metric learning framework, Fig.\ \ref{fig:flowchart}, which consists of two network branches $i.e.,$ the support branch (above) and the query branch (below). Over the support branch, PMMs are estimated for the support image(s).
In the support and query branches, two CNNs with shared weights are used as the backbone to extract features.
Let ${S} \in \mathcal{R}^{W\times H\times C}$ denote the features of the support image where $W\times H$ denotes the resolution of feature maps and $C$ the number of feature channels. The features for a query image are denoted as ${Q} \in \mathcal{R}^{W\times H\times C}$. 

Without loss of generality, the network architecture and models are illustrated for 1-shot setting, which can be extended to 5-shot setting by feeding five support images to the PMMs to estimate prototypes.

\subsection{Prototype Mixture Models}
\begin{figure}[t]
\centering
\includegraphics[width=1\columnwidth]{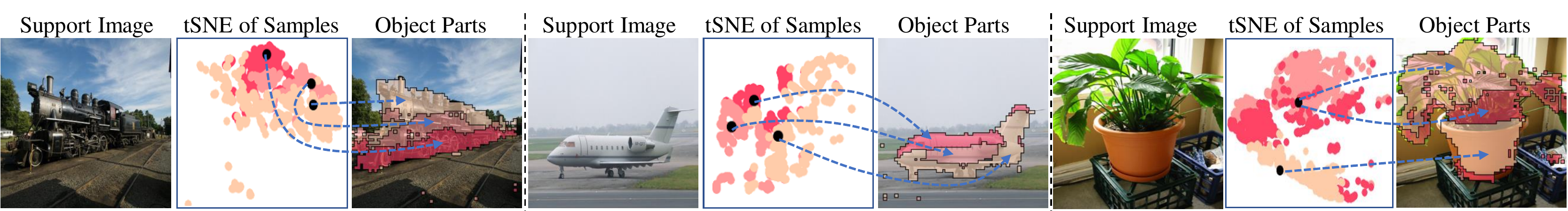}
\caption{Foreground sample distribution of support images. The black points on tSNE maps denote positive prototypes correlated to object parts. (Best viewed in color)}
\label{fig:Tsne}
\vspace{-0.2cm}
\end{figure}

During training, features $S \in \mathcal{R}^{W\times H\times C}$ for the support image are considered as a sample set with $W\times H$ $C$-dimensional samples. $S$ is spatially partitioned into foreground samples $S^+$ and background samples $S^-$, where $S^+$ corresponds to feature vectors within the mask of the support image and $S^-$ those outside the mask. 
$S^+$ is used to learn foreground PMMs$^+$ corresponding to object parts, Fig.\ \ref{fig:Tsne}, and $S^-$ to learn background PMMs$^-$. Without loss generality, the models and learning procedure are defined for PMMs, which represent either PMMs$^+$ or PMMs$^-$.

\textbf{Models.} PMMs are defined as a probability mixture model which linearly combine probabilities from base distributions, as
\begin{align}
   p(s_i|\theta) = \sum_{k=1}^K w_k p_k(s_i|\theta)
\end{align}
where $w_k$ denotes the mixing weights satisfying $0 \leq w_k \leq 1$ and $ \sum_{k=1}^K w_k =1$. $\theta$ denotes the model parameters which are learned when estimating PMMs. $s_i \in S $ denotes the $i^{th}$ feature sample and $p_k(s_i|\theta)$ denotes the $k^{th}$ base model, which is a probability model based on a Kernel distance function, as 
\begin{align}
   p_k(s_i|\theta) = \beta(\theta) e^{Kernel(s_i, \mu_k)},
\end{align}
where $\beta(\theta)$ is the normalization constant. $\mu_k \in \theta$ is one of the parameter. 
For the Gaussian mixture models (GMMs) with fixed co-variance, the Kernel function is a radial basis function (RBF), $Kernel(s_i, \mu_k) = -||(s_i - \mu_k)||_2^2$.
For the von Missies-Fisher (VMF) model~\cite{VMF2005}, the kernel function is defined as a cosine distance function, as $Kernel(s_i, \mu_k) = \frac{\mu_k^T s_i}{||\mu_k||_2 ||s_i||_2}$,
%
where $\mu_k$ is the mean vector of the $k^{th}$ model. 
Considering the metric learning framework used, the vector distance function is more appropriate in our approach, as is validated in experiments.
Based on the vector distance, PMMs are defined as
\begin{align}
   p_k(s_i|\theta) = \beta_c(\kappa) e^{\kappa \mu_k^T s_i},
\end{align}
where $\theta = \left\{ \mu, \kappa \right\}$. $\beta_c(\kappa) = \frac{\kappa^{c/2-1}}{(2\pi)^{c/2}I_{c/2-1}(\kappa)}$ is the normalization coefficient, and $I_\nu(\cdot)$ denotes the Bessel function. $\kappa$ denotes the concentration parameter, which is empirically set as $\kappa=20$ in experiments.

\textbf{Model Learning.} PMMs are estimated using the EM algorithm which includes iterative E-steps and M-steps. 
In each E-step, given model parameters and sample features extracted, we calculate the expectation of the sample $s_i$ as
\begin{equation}
\label{EStep}
   E_{ik} = \frac{p_k(s_i|\theta)}{\sum_{k=1}^K p_k (s_i \theta)} =
   \frac{e^{\kappa \mu_k^T s_i}}{\sum_{k=1}^K e^{\kappa \mu_k^T s_i}}.
\end{equation}
In each M-step, the expectation is used to update the mean vectors of PMMs, as
\begin{align}\label{MStep}
    \mu_k = \frac{\sum_{i=1}^N E_{ik}s_i}{\sum_{i=1}^N E_{ik}},
\end{align}
where $N=W\times H$ denotes the number of samples. 

After model learning, the mean vectors $\mu^+=\{\mu^+_{k}, k=1,...,K\}$  and $\mu^-=\{\mu^-_{k}, k=1,...,K\}$  are used as prototype vectors to extract convolutions features for the query image. The mixture coefficient $w_k$ is ignored so that each prototype vectors have same importance for semantic segmentation. Obviously, each prototype vector is the mean of a cluster of samples. Such a prototype vector can represent a region around an object part in the original image for the reception field effect, Fig.\ \ref{fig:Tsne}. 

\begin{figure}[t]
\centering
\includegraphics[width=0.9\columnwidth]{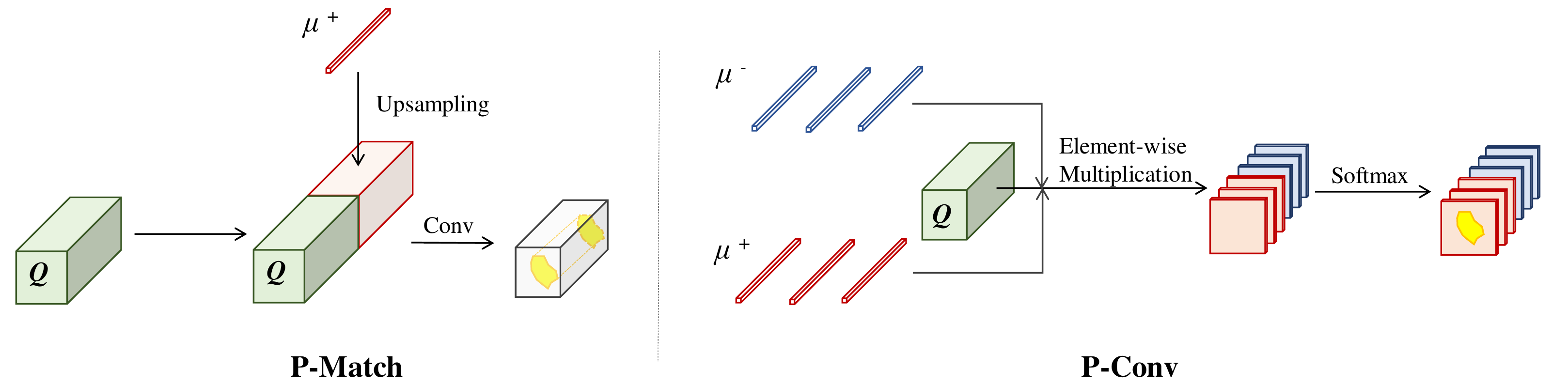}
\caption{Illustration of P-Match and P-Conv opeartions for query feature ($Q$) activation.}
\label{fig:segmentation}
\vspace{-0.2cm}
\end{figure}

\begin{algorithm}[t]
\label{alg:onlineEM}
\caption{Learning PMMs for few-shot segmentation}
\renewcommand{\algorithmicrequire}{\textbf{Input:}}  
\renewcommand{\algorithmicensure}{\textbf{Output:}}  
\begin{algorithmic}
\REQUIRE ~~\\ 
Support images, mask $M_s$ for each support image, query image;
\ENSURE ~~\\ 
Network parameters $\alpha$, prototypes $\mu^+$ and $\mu^-$ for each support image;

\FOR{(each support image)}
\STATE \textbf{Calculate} support and query features $S$ and $Q$ by forward propagation with $\theta$;\
\STATE \textbf{Partition} $S$ into $S^+$ and $S^-$ according to $M_s$;\
\STATE \textbf{Learn} PMMs;\
\STATE\quad\quad Estimate $\mu^+$ upon $S^+$ by iterative EM steps defined by  Eqs. 4 and 5;\
\STATE\quad\quad Estimate $\mu^-$ upon $S^-$ by iterative EM steps defined by  Eqs. 4 and 5;\
\STATE \textbf{Activate} $Q$ using P-Match and P-Conv defined on $\mu^+$ and $\mu^-$, Fig.\ \ref{fig:segmentation};\
\STATE \textbf{Predict} a segmentation mask and calculate the segmentation loss;\
\STATE \textbf{Update} $\alpha$ to minimize the cross-entropy loss at the query branch, Fig.\ \ref{fig:flowchart}.\
\ENDFOR
\end{algorithmic}
\end{algorithm}

\subsection{Few-shot Segmentation}
During inference, the learned prototype vectors $\mu^+=\{\mu^+_{k}, k=1,...,K\}$  and $\mu^-=\{\mu^-_{k}, k=1,...,K\}$ are duplexed to activate query features for semantic segmentation, Fig.\ \ref{fig:flowchart}. 

\textbf{PMMs as Representation (P-Match)}. 
Each positive prototype vector squeezes representative information about an object part and all prototypes incorporate representative information about the complete object extent. 
Therefore, prototype vectors can be used to match and activate the query features $Q$, as
\begin{equation}
  Q' = \text{P-Match}(\mu_k^+, Q), k=1,...,K,
\end{equation}
where P-Match refers to an activation operation consists of prototype upsampling, feature concatenation, and semantic segmentation using convolution, Fig.\ \ref{fig:segmentation}.
The convolution operation on concatenated features implements a channel-wise comparison, which activates feature channels related to foreground while suppressing those associated with backgrounds. With P-Match, semantic information about the extent of the complete object is incorporated into the query features for semantic segmentation.

\textbf{PMMs as Classifiers (P-Conv)}. 
On the other hand, each prototype vector incorporating discriminative information across feature channels can be seen as a classifier, which produces probability maps $M_k=\{M_k^+,M_k^-\}$ using the P-Conv operation, as
\begin{equation}
\begin{split}
  M_k &= \text{P-Conv}(\mu_k^+,\mu_k^-,Q), k=1,...,K.\\
  \end{split}
\end{equation}
As shown in Fig.\ \ref{fig:segmentation}, P-Conv first multiplies each prototype vector with the query feature $Q$ in an element-wise manner. The output maps are then converted to probability maps $M_k$ by applying Softmax across channels.

After P-Conv, the produced probability maps $M_k^+, k=1,...,K$ and $M_k^-, k=1,...,K$ are respectively summarized to two probability maps, as 
\begin{equation}
\begin{split}
  M_p^+ = \sum_k M_k^+,\\
  M_p^- = \sum_k M_k^-,
 \end{split}
\end{equation}
which are further concatenated with the query features to activate objects of interest, as 
\begin{equation}
  Q'' = M_p^+ \oplus M_p^- \oplus  Q',
\end{equation}
where $\oplus$ denotes the concatenation operation. 

After the P-Match and P-Conv operations, the semantic information across channels and discriminative information related to object parts are collected from the support feature $S$ to activate the query feature $Q$. in a dense comparison manner~\cite{CaNet}. The activated query features $Q''$ are further enhanced with Atrous Spatial Pyramid Pooling (ASPP) and fed to a convolutional module to predict the segmentation mask, Fig.\ \ref{fig:flowchart}. 

\begin{figure}[t]
\centering
\includegraphics[width=1.0\columnwidth]{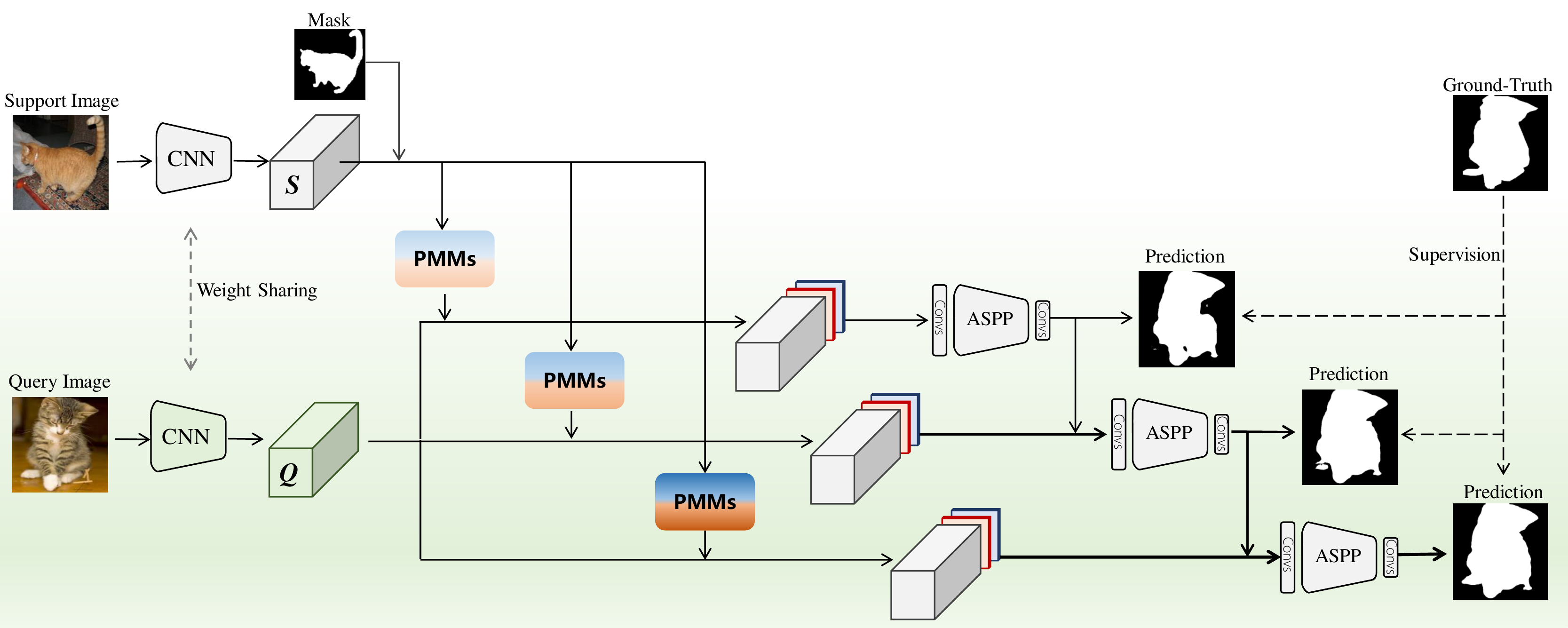}
\caption{Network architecture of residual prototype mixture models (RPMMS).}
\label{fig:residual}
\vspace{-0.2cm}
\end{figure}

\textbf{Segmentation Model Learning}.
The segmentation model is implemented as an end-to-end network, Fig.\ \ref{fig:flowchart}.
The learning procedure for the segmentation model is described in Algorithm 1. In the feed forward procedure, the support features are partitioned into backgrounds and foreground sample sets $S^+$ and $S^-$ according to the ground-truth mask. 
PMMs are learned on $S^+$ and $S^-$ and the learned prototype vectors $\mu^+$ and $\mu^-$ are leveraged to activate query features to predict segmentation mask of the query image. In the back-propagation procedure, the network parameters $\theta$ are updated to optimize the segmentation loss at the query branch. With multiple training iterations, rich feature representation about diverse object parts is absorbed into the backbone network. 
During the inference procedure, the learned feature representation together with PMMs of the support image(s) is used to segment the query image. 

\subsection{Residual Prototype Mixture Models}
To further enhance the model representative capacity, we implement model ensemble by stacking multiple PMMs, Fig.\ \ref{fig:residual}. Stacked PMMs, termed residual PMMs (RPMMs), leverage the residual from the previous query branch to supervise the next query branch for fine-grained segmentation. 
This is computationally easier as it pursuits the minimization of residuals between branches rather than struggling to combine multiple models to fit a ground-truth mask. RPMMs not only further improve the performance but also defines a new model ensemble strategy. This incorporates the advantages of model residual learning, which is inspired by the idea of side-output residual~\cite{SRN17,RSRN17} but has the essential difference to handle models instead of features. This is also different from the ensemble of experts~\cite{FWB-ICCV2019}, which generates an ensemble of the support features guided by the gradient of loss.

\section{Experiments}

\subsection{Experimental settings}

\textbf{Implementation Details.}
Our approach utilizes CANet~\cite{CaNet} without iterative optimization as the baseline, which uses VGG16 or ResNet50 as backbone CNN for feature extraction. During training, four data augmentation strategies including normalization, horizontal flipping, random cropping and random resizing are used~\cite{CaNet}. Our approach is implemented upon the PyTorch 1.0 and run on Nvidia 2080Ti GPUs. 
The EM algorithm iterates 10 rounds to calculate PMMs for each image.
The network with a cross-entropy loss is optimized by SGD with the initial learning of 0.0035 and the momentum of 0.9 for 200,000 iterations with 8 pairs of support-query images per batch. The learning rate reduces following the ``poly" policy defined in DeepLab~\cite{DeepLabV2}. For each training step, the categories in the train split are randomly selected and then the support-query pairs are randomly sampled in the selected categories.

\textbf{Datasets.} 
We evaluate our model on Pascal-5$^i$ and COCO-20$^i$.
Pascal-5$^i$ is a dataset specified for few-shot semantic segmentation in OSLSM~\cite{OSLSM}, which consists of the Pascal VOC 2012 dataset with extra annotations from extended SDS~\cite{SDS11}. 20 object categories are partitioned into four splits with three for training and one for testing. At test time, 1000 support-query pairs were randomly sampled in the test split~\cite{CaNet}. Following FWB~\cite{FWB-ICCV2019}, we create COCO-20$^i$ from MSCOCO 2017 dataset. The 80 classes are divided into 4 splits and each contains 20 classes and the $val$ dataset is used for performance evaluation. The other setting is the same as that in Pascal-5$^i$.

\textbf{Evaluation Metric.}
Mean intersection over-union (mIoU) which is defined as the mean IoUs of all image categories was employed as the metric for performance evaluation. For each category, the IoU is calculated by IoU=$\frac{\text{TP}}{\text{TP+FP+FN}}$, where TP, FP and FN respectively denote the number of true positive, false positive and false negative pixels of the predicted segmentation masks. 

\begin{figure}[t]
\centering
\includegraphics[width=1\linewidth]{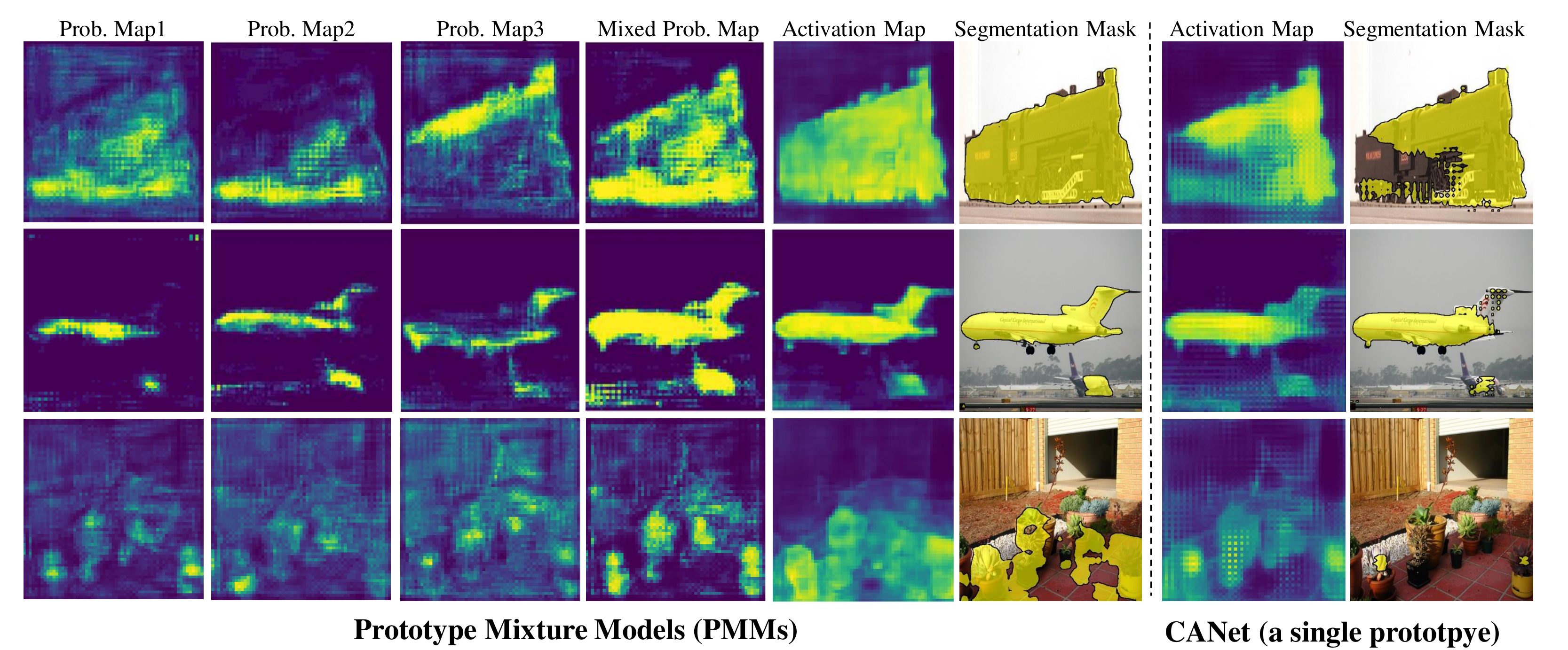}
\caption{Activation maps by PMMs and CANet~\cite{CaNet}. PMMs produce multiple probability maps and fuse them to a mixed map, which facilities activating and segmenting complete object extent (first two rows) or multiple objects (last row). CANet that uses a single prototype to segment object tends to miss object parts. (Best viewed in color)}
\label{fig:prob_map}
\vspace{-0.2cm}
\end{figure}

\subsection {Model Analysis}

In Fig.\ \ref{fig:prob_map}, we visualize probability maps produced by positive prototypes of PMMs. We also visualize and compare the activation maps and segmentation masks produced by PMMs and CANet. PMMs produce multiple probability maps and fuse them to a mixed probability map, which facilities activating complete object extent (first two rows).
The advantage in terms of representation capacity is that PMMs perform better than CANet when segmenting multiple objects within the same image (last row). 
By comparison, CANet using a single prototype to activate object tends to miss object parts or whole objects. The probability maps produced by PMMs validate our idea, $i.e.$, prototypes correlated to multiple regions and alleviate semantic ambiguity.

\begin{figure}[!t]
\centering
\includegraphics[width=1\linewidth]{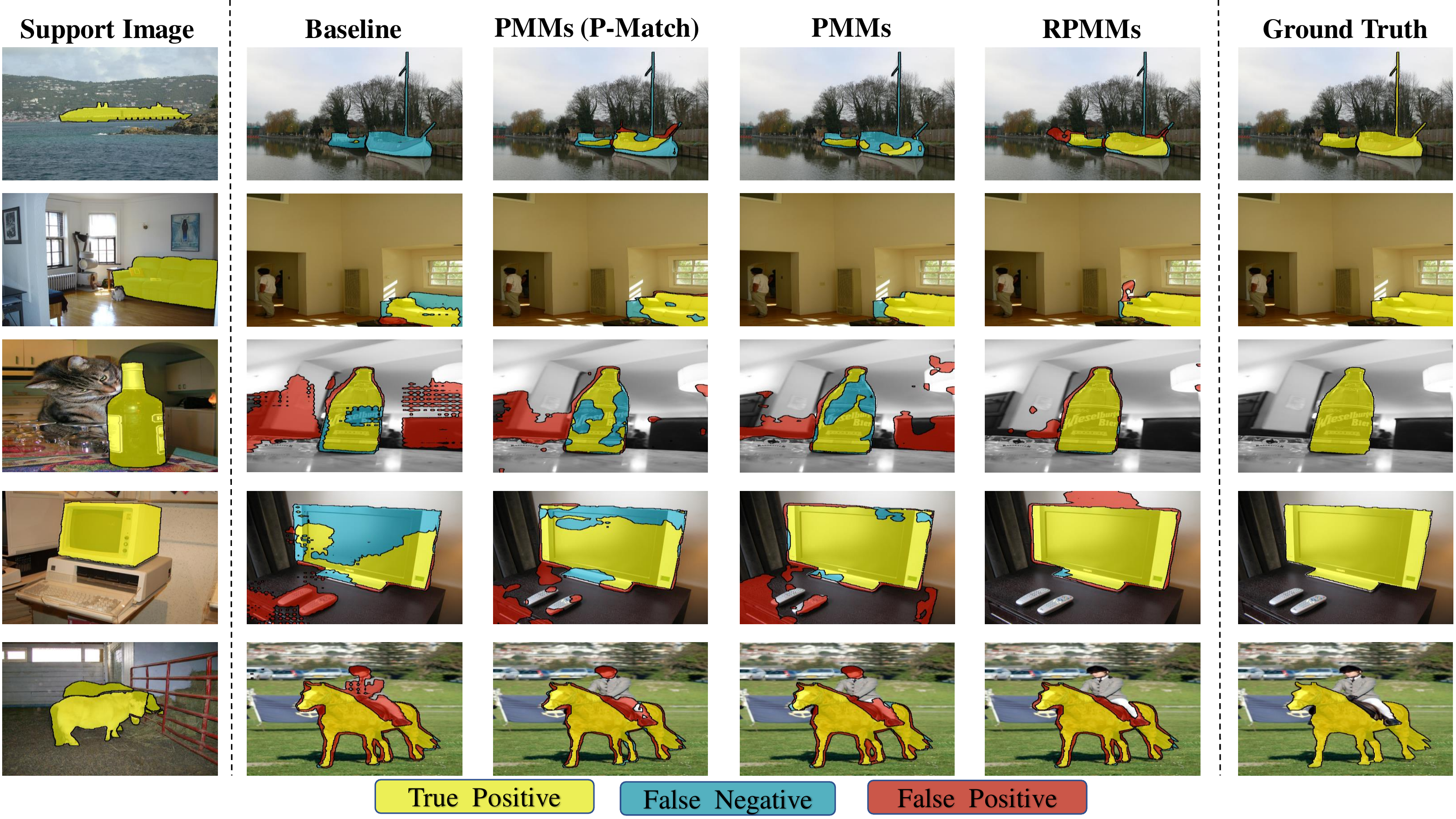}
\caption{Semantic segmentation results. 
`Baseline' refers to the CANet method~\cite{CaNet} without iterative optimization. (Best viewed in color)}
\label{fig:ablation}
\vspace{-0.2cm}
\end{figure}

In Fig.\ \ref{fig:ablation}, we compare the segmentation results by the baseline method and the proposed modules. The segmentation results show that PMMs$^+$(P-Match) can improve the recall rate by segmenting more target pixels.
%
By introducing background prototypes, PMMs reduce the false positive pixels, which validates that the background mixture models can improve the discriminative capability of the model. 
RPMMs further improve the segmentation results by refining object boundaries about hard pixels.

\setlength{\tabcolsep}{4pt}
\begin{table}[t]
\begin{center}
\caption{Ablation study. `Mean' denotes mean mIoU on Pascal-5$^i$ with PMMs using three prototypes. The first row is the baseline method without using the PMMs or the RPMMs method.
}
\label{table:PMM_performance}
\begin{tabular}{lllllllll}
\hline\noalign{\smallskip}
PMMs$^+$(P-Match) & PMMs(P-Match \& P-Conv) & RPMMs & Mean \\
\noalign{\smallskip}
\hline
\noalign{\smallskip}
            &           &           & 51.93\\
 \checkmark &           &           & 54.63\\
 \checkmark & \checkmark&           & 55.27\\
 \checkmark &\checkmark &\checkmark & 56.34\\
\hline

\end{tabular}
\end{center}
\end{table}
\setlength{\tabcolsep}{1.4pt}


\setlength{\tabcolsep}{4pt}
\begin{table}[t]
\begin{center}
\caption{Performance (mIoU\%) on prototype number $K$.}
\label{table:k}
\begin{tabular}{lllllll}
\hline\noalign{\smallskip}
$K$ & Pascal-$5^0$ & Pascal-$5^1$ & Pascal-$5^2$ & Pascal-$5^3$ & Mean \\
\noalign{\smallskip}
\hline
\noalign{\smallskip}
 1& 49.38 & 66.42 & 51.29 & 47.68 & 53.69\\
 2& 50.85 & 66.65 & \bf51.89 & 48.25 & 54.41\\
 3& 51.88 & 66.72 & 51.14 & \bf48.80 & \bf54.63\\
 4& \bf51.89 &\bf 66.96 & 51.36 & 47.91 & 54.53\\
 5& 50.76 & 66.89 & 50.76 & 47.94 & 54.09\\
\hline
\end{tabular}
\end{center}
\end{table}
\setlength{\tabcolsep}{1.4pt}

\setlength{\tabcolsep}{4pt}
\begin{table}
\begin{center}
\caption{Performance comparison of Kernel functions.}
\label{table:kernel}
\begin{tabular}{llllll}
\hline\noalign{\smallskip}
Kernal & Pascal-$5^0$ & Pascal-$5^1$ & Pascal-$5^2$ & Pascal-$5^3$ & Mean \\
\noalign{\smallskip}
\hline
\noalign{\smallskip}
 Gaussian & 50.94 & 66.70 & 50.59 & 47.91 & 54.04 \\
 {\bf VMF} & {\bf 51.88} & {\bf 66.72} & {\bf 51.14} & {\bf 48.80} & {\bf 54.63}\\
 \hline
\end{tabular}
\end{center}
\end{table}
\setlength{\tabcolsep}{1.4pt}

\subsection{Ablation Study} 

%
\textbf{PMMs.}
In Table\ \ref{table:PMM_performance}, with P-Match modules, PMMs improve segmentation performance by 2.70\% (54.63\% vs. 51.93\%), which validates that the prototypes generated by the PMMs perform better than the prototype generated by global average pooling.
By introducing the duplex strategy, PMMs further improve the performance by 0.64\% (55.27\% vs. 54.63\%),  which validates that the probability map generated by the combination of foreground and background prototypes can suppress backgrounds and reduce false segmentation.
In total, PMMs improve the performance by 3.34\% (55.27\% vs. 51.93\%), which is a significant margin in semantic segmentation. This clearly demonstrates the superiority of the proposed PMMs over previous prototype methods. 

\textbf{RPMMs}. RPMMs further improve the performance by 1.07\% (56.34\% vs. 55.27\%), which validates the effectiveness of the residual ensemble strategy. Residual from the query prediction output of the previous branch of PMMs can be used to supervise the next branch of PMMs, enforcing the stacked PMMs to reduce errors, step by step.

\setlength{\tabcolsep}{4pt}
\begin{table}[t]
\begin{center}
\caption{Performance of 1-way 1-shot semantic segmentation on Pascal-5$^i$. CANet reports multi-scale test performance. The single-scale test performance is obtained from github.com/icoz69/CaNet/issues/4 for a fair comparison.}
\label{table:1shot}
\begin{tabular}{lllllll}
\hline\noalign{\smallskip}
Backbone & Method & Pascal-$5^0$ & Pascal-$5^1$ & Pascal-$5^2$ & Pascal-$5^3$ & Mean \\
\noalign{\smallskip}
\hline
\noalign{\smallskip}
\multirow{5}{*}{VGG16} & OSLSM~\cite{OSLSM} & 33.60 & 55.30 & 40.90 & 33.50 & 40.80 \\
  & co-FCN~\cite{co-FCN}                    & 36.70 & 50.60 & 44.90 &32.40 & 41.10\\
  & SG-One~\cite{SG-One}                    & 40.20 & 58.40 & 48.40 &38.40 & 46.30\\
  & PANet~\cite{PANet}                      & 42.30 & 58.00 & 51.10 &41.20 & 48.10\\
  & FWB~\cite{FWB-ICCV2019}                 & 47.04 & 59.64 & {\bf 52.51} & 48.27 & 51.90\\
   & \bf RPMMs (ours)                       & \bf 47.14& \bf 65.82 & 50.57 &\bf 48.54 & \bf53.02\\
\hline
\multirow{2}{*}{Resnet50} & CANet~\cite{CaNet}  & 49.56 & 64.97 & 49.83 & {\bf51.49} & 53.96 \\
  & {\bf  PMMs (ours)}                         & 51.98 & \bf 67.54 & 51.54 & 49.81 & 55.22\\
  & {\bf RPMMs (ours)}                         & \bf 55.15 & 66.91 & \bf 52.61 & 50.68 & {\bf 56.34}\\
\hline
\vspace{-0.2cm}
\end{tabular}
\end{center}
\end{table}
\setlength{\tabcolsep}{1.4pt}

\setlength{\tabcolsep}{4pt}
\begin{table}[t]
\begin{center}
\caption{Performance of 1-way 5-shot semantic segmentation on Pascal-5$^i$.}
\label{table:5shot}
\begin{tabular}{lllllll}
\hline\noalign{\smallskip}
Backbone & Method & Pascal-$5^0$ & Pascal-$5^1$ & Pascal-$5^2$ & Pascal-$5^3$ & Mean \\
\noalign{\smallskip}
\hline
\noalign{\smallskip}
\multirow{5}{*}{VGG16} & OSLSM~\cite{OSLSM} & 35.90 & 58.10 & 42.70 & 39.10 & 43.95 \\
  & SG-One~\cite{SG-One}                    & 41.90 & 58.60 & 48.60 &39.40 & 47.10  \\
  & FWB~\cite{FWB-ICCV2019}                 & 50.87 & 62.86 & 56.48 &\bf 50.09 & 55.08 \\
  & \bf PANet~\cite{PANet}                  & \bf51.80 & 64.60 & \bf59.80 & 46.05 & \bf55.70 \\
  & {RPMMs (ours)}                          & 50.00 & \bf66.46 & 51.94 & 47.64 & 54.01\\
\hline
\multirow{2}{*}{Resnet50} 
& CANet~\cite{CaNet}                       &- &- &- &- & 55.80 \\
& {\bf PMMs (ours)}                        &55.03  &\bf68.22  &52.89  &\bf51.11  &56.81  \\
& \bf RPMMs (ours)                         &\bf56.28 &67.34 &\bf54.52 &51.00 &\bf57.30\\
\hline
\vspace{-0.2cm}
\end{tabular}
\end{center}
\end{table}
\setlength{\tabcolsep}{1.4pt}

\setlength{\tabcolsep}{4pt}
\begin{table}[!t]
\begin{center}
\caption{Performance of 1-shot and 5-shot semantic segmentation on MS COCO. FWB uses the ResNet101 backbone while other approaches use the ResNet50 backbone.}
\label{table:coco-res}
\begin{tabular}{lllllll}
\hline\noalign{\smallskip}
Settings & Method & COCO-$20^0$ & COCO-$20^1$ & COCO-$20^2$ & COCO-$20^3$ & Mean \\
\noalign{\smallskip}
\hline
\multirow{5}{*}{1-shot} 
& PANet~\cite{PANet}                     &- &- &- &- & 20.90\\
& FWB~\cite{FWB-ICCV2019}                &16.98 &17.98 &20.96 &28.85 & 21.19\\
& Baseline                               &25.08 &30.25 &24.45 &24.67 &26.11\\
& {\bf PMMs (ours)}                      &29.28  &34.81  &27.08  &27.27  &29.61  \\
& {\bf RPMMs (ours)}                      &\bf29.53  &\bf36.82  &\bf28.94  &\bf27.02  &\bf30.58  \\
\hline
\multirow{5}{*}{5-shot} 
& FWB ~\cite{FWB-ICCV2019}              &19.13&21.46 &23.93 &30.08 & 23.65 \\
& PANet~\cite{PANet}                     &- &- &- &- & 29.70 \\
& Baseline                               &25.95 &32.38 &26.11 &26.98 &27.86\\
& {\bf PMMs (ours)}                      &33.00  &40.55  &30.29  &33.27  &34.28  \\
& {\bf RPMMs (ours)}                     &\bf33.82  &\bf41.96  &\bf32.99  &\bf33.33  &\bf35.52  \\
\hline
\vspace{-0.2cm}
\end{tabular}
\end{center}
\end{table}
\setlength{\tabcolsep}{1.4pt}   

\textbf{Number of Prototypes.} In Table \ref{table:k}, ablation study is carried out to determine the number of prototypes using PMMs$^+$ with P-Match. $K=2$ significantly outperforms $K=1$, which validates the plausibility of introducing mixture prototypes. 
The best Pascal-5$^i$ performance occurs at $K=2,3,4$. When $K=3$ the best mean performance is obtained. When $K=4,5$, the performance slightly decreases. One reason lies in that the PMMs are estimated on a single support image, which includes limited numbers of samples. The increase of $K$ substantially decreases the samples of each prototype and increases the risk of over-fitting. 

\textbf{Kernel Functions.} In Table~\ref{table:kernel}, we compare the Gaussian and VMF kernels for sample distance calculation when estimating PMMs. The better results from VMF kernel show that the cosine similarity defined by VMF kernel is preferable.

\textbf{Inference Speed.} 
The size of PMMs model is 19.5M, which is slightly larger than that of the baseline CANet~\cite{CaNet} (19M) but much smaller than that of OSLSM~\cite{OSLSM} (272.6M). Because the prototypes are $1\times 1\times C$ dimensional vectors, they do not significantly increase the model size or computational cost. In one shot setting, with $K=3$, our inference speed on single 2080Ti GPU is 26 FPS, which is slightly lower than that of CANet (29 FPS). With RPMMs the speed decreases to 20 FPS while the model size (19.6M) does not significantly increase.

\subsection{Performance}
\textbf{PASCAL-5$^i$.} 
In Table \ref{table:1shot} and Table \ref{table:5shot}, PMMs and RPMMs are compared with the state-of-the-art methods. They outperform state-of-the-art methods in both 1-shot and 5-shot settings. With the 1-shot setting and a Resnet50 backbone, RPMMs achieve 2.38\% (56.34\% vs. 53.96\%) performance improvement over the state-of-the-art, which is a significant margin. 

With the 5-shot setting and a Resnet50 backbone, RPMMs achieve 1.50\% (57.30\% vs. 55.80\%) performance improvement over the state-of-the-art, which is also significant. With the VGG16 backbone, our approach is comparable with the state-of-the-arts. Note that the PANet and FWB used additional $k$-shot fusion strategy while we do not use any post-processing strategy to fuse the predicted results from five shots.
%

\textbf{MS COCO.} Table~\ref{table:coco-res} displays the evaluation results on MS COCO dataset following the evaluation metric on COCO-$20^i$~\cite{FWB-ICCV2019}. Baseline is achieved by running CANet without iterative optimization. PMMs and RPMMs again outperform state-of-the-art methods in both 1-shot and 5-shot settings. 
For the 1-shot setting, RPMMs improves the baseline by 4.47\%, respectively outperforms the PANet and FWB methods by 9.68\% and 9.39\%. 

For the 5-shot setting, it improves the baseline by 7.66\%, and respectively outperforms the PANet and FWB by 5.82\% and 11.87\%, which are large margins for the challenging few-shot segmentation problem. Compared to PASCAL VOC, MS COCO has more categories and images for training, which facilities learning richer representation related to various object parts and backgrounds. Thereby, the improvement on MS COCO is larger than that on Pascal VOC.

\section{Conclusion}
We proposed prototype mixture models (PMMs), which correlate diverse image regions with multiple prototypes to solve the semantic ambiguity problem. During training, PMMs incorporate rich channel-wised and spatial semantics from limited support images. During inference, PMMs are matched with query features in a duplex manner to perform accurate semantic segmentation. On the large-scale MS COCO dataset, PMMs improved the performance of few-shot segmentation, in striking contrast with state-of-the-art approaches. As a general method to capture the diverse semantics of object parts given few support examples, PMMs provide a fresh insight for the few-shot learning problem.

\textbf{Acknowledgement:} This work was supported in part by the National Natural Science Foundation of China (NSFC) under Grant 61836012, 61671427, and 61771447. 

\clearpage

\bibliographystyle{splncs}
\bibliography{main}
\end{document}